\documentclass[10pt,twocolumn,letterpaper]{article}

\usepackage{cvpr}      

\usepackage{graphicx}
\usepackage{comment}
\usepackage{amsmath,amssymb} 
\usepackage{color}
\usepackage{physics}
\usepackage{enumitem}

\usepackage{xspace}
\usepackage{multirow}
\usepackage{bbm}
\usepackage{booktabs}
\usepackage{multirow}

\usepackage{algorithm}
\usepackage[noend]{algpseudocode}
\usepackage{tabularx}
\usepackage{makecell}
\usepackage{microtype}
\usepackage{textcomp}
\usepackage{colortbl}
\usepackage[table]{xcolor}
\usepackage{romannum}
\usepackage{url}
\usepackage{wrapfig}

\usepackage[pagebackref,breaklinks,colorlinks]{hyperref}

\def\eg{e.g.,~} 
\newcommand\savemathcal[1]{%
  \expandafter\newsavebox\csname mc#1content\endcsname%
  \expandafter\savebox\csname mc#1content\endcsname{$\mathcal{#1}$}%
  \expandafter\newcommand\csname mc#1\endcsname{%
    \expandafter\usebox\expandafter{\csname mc#1content\endcsname}}%
}
\newcommand\altmathcal[1]{\csname mc#1\endcsname}
\savemathcal{E}
\savemathcal{F}
\savemathcal{S}
\usepackage{bm}

\newcommand{\beq}{\begin{equation}}
\newcommand{\eeq}{\end{equation}}

\newcommand{\Rdom}{\mathbb{R}^2}

\newcommand{\D}{\mathcal{D}}

\usepackage[capitalize]{cleveref}
\crefname{section}{Sec.}{Secs.}
\Crefname{section}{Section}{Sections}
\Crefname{table}{Table}{Tables}
\crefname{table}{Tab.}{Tabs.}

\def\cvprPaperID{10574} 
\def\confName{CVPR}
\def\confYear{2023}

\begin{document}
\pagenumbering{arabic}
\title{Neural Lens Modeling}
%

\author{Wenqi Xian$^{1,*}$\hspace{5mm}
Alja\v{z} Bo\v{z}i\v{c}$^2$\hspace{5mm}
Noah Snavely$^3$\hspace{5mm}
Christoph Lassner$^4$\\
Meta Reality Labs Research$^{1, 2, 4}$\\
Cornell University$^{1, 3}$\\
{\tt\small wx97@cornell.edu$^1$, aljaz@meta.com$^2$, snavely@cs.cornell.edu$^3$, classner@meta.com$^4$}
}
\maketitle

\iftoggle{cvprfinal}{%
\renewcommand*{\thefootnote}{\fnsymbol{footnote}}
\footnotetext[1]{Work done during an internship at RLR.}
\renewcommand*{\thefootnote}{\arabic{footnote}}
}{}

\begin{abstract}

Recent methods for 3D reconstruction and rendering increasingly benefit from end-to-end optimization of the entire image formation process.
%
However, this approach is currently limited: effects of the optical hardware stack and in particular lenses are hard to model in a unified way.
This limits the quality that can be achieved for camera calibration and the fidelity of the results of 3D reconstruction.
In this paper, we propose NeuroLens, a neural lens model for distortion and vignetting that can be used for point projection and ray casting and can be optimized through both operations.
This means that it can (optionally) be used to perform pre-capture calibration using classical calibration targets, and can later be used to perform calibration or refinement during 3D reconstruction, e.g., while optimizing a radiance field.
To evaluate the performance of our proposed model, we create a comprehensive dataset assembled from the Lensfun database with a multitude of lenses.
Using this and other real-world datasets, we show that the quality of our proposed lens model outperforms standard packages as well as recent approaches while being much easier to use and extend.
The model generalizes across many lens types and is trivial to integrate into existing 3D reconstruction and rendering systems.
Visit our project website at: \url{https://neural-lens.github.io}.
\vspace{-5mm}
\end{abstract}


\begin{figure}[t]
\maketitle
\begin{center}
\includegraphics[width=\linewidth]{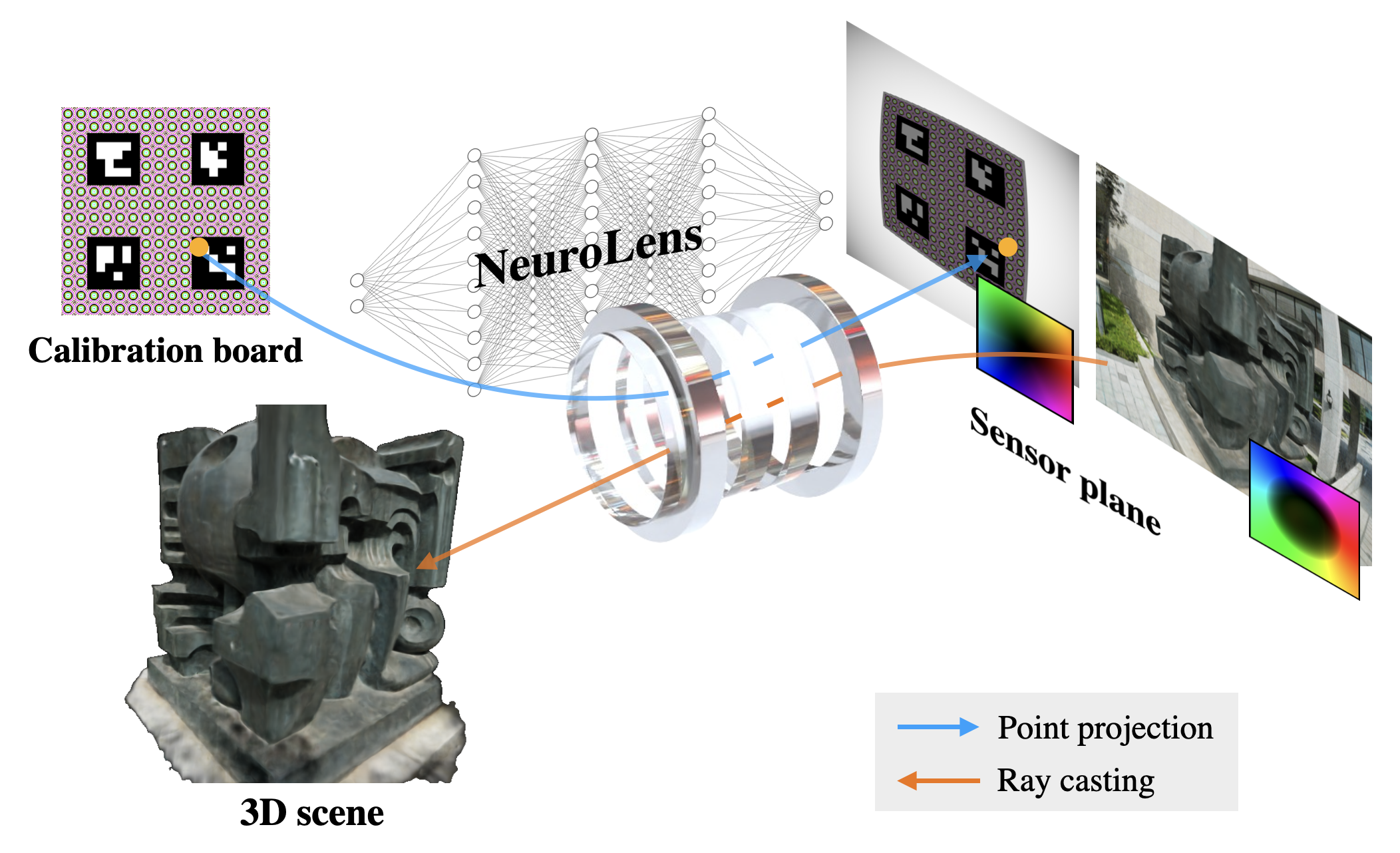}
\end{center}
\vspace{-4mm}
\caption{\textbf{Method Overview.} The optical stack leads to light ray distortion and vignetting. We show that invertible residual networks are a powerful tool to model the distortion for projection and ray casting across many lenses and in many scenarios. Additionally, we propose a novel type of calibration board (top left) that can optionally be used to improve calibration accuracy. For evaluation, we propose the `SynLens' dataset to evaluate lens models at scale.\protect\footnotemark
}
\label{fig:method}
\vspace{-5mm}
\end{figure}

\footnotetext{The approach is visualized on FisheyeNeRF recordings~\cite{jeong2021self}.}

\section{Introduction}


Camera calibration is essential for many computer vision applications: it is the crucial component mapping measurements and predictions between images and the real world.
This makes calibration a fundamental building block of 3D reconstruction and mapping applications, and 
of any system that relies on spatial computing, such as autonomous driving or augmented and virtual reality.
Whereas camera extrinsics and the parameters of a pinhole model can be easily described and optimized, this often does not hold for other parameters of an optical system and, in particular, lenses.
Yet lenses have a fundamental influence on the 
captured image through distortion and vignetting effects.

Recent results in 3D reconstruction and rendering suggest that end-to-end modeling and optimization of the image formation process leads to the highest fidelity 
scene reproductions~\cite{Lombardi:2019,mildenhall2020nerf}.
Furthermore, per-pixel gradients are readily available in this scenario and \emph{could} serve as a means to optimize a model of all components of the optical stack to improve reconstruction quality.
However, modeling and optimizing lens parameters in full generality and also \emph{differentiably}
is hard: camera lenses come in all kinds of forms and shapes (\eg pinhole, fisheye, catadioptric) with quite different optical effects.
%
%
%
%
%


%
So how can we create a flexible and general and differentiable lens model with enough parameters to approximate any plausible distortion? 
In classical parametric models, the internals of the camera are assumed to follow a model with a limited number of parameters (usually a polynomial approximation).
These approaches work well when the distortion is close to the approximated function, but cannot generalize beyond that specific function class. 
%
On the other hand, non-parametric models that associate each pixel with a 3D ray have also been explored.
These models are designed to model any type of lens, but tend to require dense keypoint measurements due to over-parameterization.
%
Hence, we aim to find models with some level of regularization to prevent such issues, 
without unnecessarily constraining the complexity of the distortion function.
Our key insight is to use an invertible neural network (INN) to model ray distortion, combined with standard camera intrinsics and extrinsics.
This means that we model the camera lens as a mapping of two vector fields using a diffeomorphism (\ie, a bijective mapping where both the mapping and its inverse are differentiable), represented by an INN.
%
This approach usefully leverages the invertibility constraints provided by INNs to model the underlying physics of the camera lens.

Our lens model has several advantages.
Its formulation makes it easy to differentiate point projection and ray casting operations in deep learning frameworks and it can be integrated into \emph{any} end-to-end differentiable pipeline, with an inductive bias that serves as a useful regularizer for lens models.
%
%
It is flexible: we can scale the model parameters to adapt to different kinds of lenses.
using gradient-based methods for point projection as well as ray casting.
This makes our model applicable to pattern-based camera calibration as well as to dense reconstruction where camera parameter refinement is desired.
In the case of (optional) marker-based calibration, we suggest to use an end-to-end optimized marker board and keypoint detector.
The proposed marker board outperforms several other alternatives in our experiments, and can easily be adjusted to be particularly robust to distortions of different sensor and lens types.

It is currently impossible to 
evaluate lens models 
at scale in a standardized way: large-scale camera lens benchmarks including ground truth data simply do not exist.
%
We propose to address this issue by generating a synthetic dataset, called SynLens, consisting of more than 400 different lens profiles from the open-source Lensfun database.
%
To create SynLens,
we simulate distortion and vignetting and (optionally) keypoint extraction noise using real lens characteristics to account for a wide variety of lenses and cameras.
%

We provide qualitative and quantitative 
comparisons with prior works and show that our method produces more accurate results in a wide range of settings, including
pre-calibration using marker boards, fine-tuning camera models during 3D reconstruction, and using quantitative evaluation on the proposed SynLens dataset.
%
%
We show that our model achieves subpixel 
accuracy even with just a few keypoints and is robust to noisy keypoint detections.
%
%
%
%
The proposed method is conceptually simple and flexible, yet achieves state-of-the-art results on calibration problems.
We attribute this success to the insight that an INN provides a useful inductive bias for lens modeling and validate this design choise via ablations on ResNet-based models.
To summarize, we claim the following contributions:
\begin{itemize}[noitemsep, leftmargin=14pt]
  \item A novel formulation and analysis of an invertible ResNet-based lens distortion model that generalizes across many lens types, is easy to implement and extend;
 \item A new way to jointly optimize marker and keypoint detectors to increase the robustness of pattern-based calibration;
 \item A large-scale camera lens benchmark for evaluating the performance of marker detection and camera calibration;
 \item Integration of the proposed method into a neural rendering pipeline as an example of purely photometric calibration. 
\end{itemize}
\section{Related Work}
\paragraph{Existing camera calibration methods.}
Many 3D computer vision methods 
assume that lens distortion is radially symmetric around the center of the image. Various camera models such as the radial~\cite{duane1971close} (bicubic~\cite{kilpela1981compensation}), division~\cite{fitzgibbon2001simultaneous}, FOV models~\cite{devernay2001straight}, and rational model~\cite{claus2005rational} are used to simulate such radially symmetric distortion. Numerous
calibration toolboxes and pipelines ~\cite{zhang2000flexible, sturm2004generic, swaninathan2003perspective} have been developed and integrated to OpenCV~\cite{bradski2000opencv}. Recently, BabelCalib~\cite{lochman2021babelcalib} proposed a robust optimization strategy for parametric models. 
However, parametric models are only approximate models of real lenses;
in practice, the real distortion includes effects caused by complex lens systems (which lead to combinations of different types of distortions) determined by the camera geometry and by the (not perfectly planar) shape of the lens~\cite{tang2017precision}.    
%
%
%

When calibrating a camera system with an unknown lens it is difficult to decide in advance which particular model fits the real type of camera projection best.
To avoid having to choose, 
one can instead use a single generic model to approximate most common types of projection.
A generic camera model~\cite{grossberg2001general, ramalingam2005towards, hartley2007parameter, nister2005non, camposeco2015non} associates each pixel with a 3D ray. 
%
%
These methods are designed for generality and flexibility and introduce an extreme number of parameters. 
%
%
In practice, classical sparse calibration patterns do not provide enough measurements for such generic models. \cite{dunne2010efficient, bergamasco2017parameter}~uses these models to obtain dense matches using displays that can encode their pixel positions or interpolate between sparse features. However, interpolation leads to inaccurate and sub-optimal performance. Therefore, models with lower calibration data requirements have been proposed~\cite{ramalingam2016unifying}. Recently, Sch\"ops \etal~\cite{schops2020having} extends~\cite{ramalingam2016unifying} with a new calibration patterns and detectors to improve the  calibration accuracy for generic cameras. \cite{pan2022camera}~replaces the explicit parametric model with a regularization term that forces the underlying distortion map to be smooth.

\medskip \noindent \textbf{Neural network--based camera calibration.}
Several prior works treat the optical components of displays and cameras as differentiable layers (neural network layers) that can be trained jointly with the computational blocks of an imaging/display system~\cite{sitzmann2018end,tseng2019hyperparameter,heide2014flexisp}.
Other works estimate camera parameters from single image observations using CNNs~\cite{bogdan2018deepcalib, xian2019uprightnet}.
For multi-view, joint optimization of camera parameters and neural scene representations, 
representative works include BARF~\cite{lin2021barf}, NeRF$--$~\cite{wang2021nerf}, Self-Calibrating Neural Radiance Fields~\cite{jeong2021self} and the point-based neural rendering pipeline of R{\"u}ckert et al.~\cite{ruckert2022adop}

\medskip \noindent \textbf{Learned markers and keypoint detectors.}
Lens models can either be optimized during 3D reconstruction or in a separate calibration stage that uses keypoint positions corresponding to a known 3D structure.
Many calibration packages use a checkerboard pattern~\cite{opencv_library} due to its simplicity and to be able to 
utilize line fitting to increase corner detection accuracy.
Sch\"ops et al.~\cite{schops2020having} propose 
a star-based pattern similar to Siemens stars~\cite{siemensstar} 
to increase the amount of gradient information available.
They use AprilTags~\cite{olson2011apriltag} to initialize their point search, while we use ArUco tags~\cite{aruco1Garrido-Jurado2016,aruco2Romero-Ramirez2018} in a similar way on our proposed marker board.

However, all these boards are manually designed.
In contrast, \cite{opencvrandpattern}~uses a random pattern optimized to produce strong feature responses for keypoint detectors.
This leads to significantly more points (on the order of thousands), albeit with lower detection accuracy.
Hu et al.~\cite{deepcharuco} propose to use a deep-learning based detector.
%
%
Grinchuk et al.~\cite{learnable_markers} propose to use a learning-based approach for creating markers by generating binary codes and rendering them on distorted and transformed image patches.
Peace et al.~\cite{peace2020e2etag,deepformabletag2021} use end-to-end trainable systems for marker detection, but focus on fiducial-like markers with a unique marker identification.
These systems usually require larger markers with a unique identifier to enable direct estimation of camera pose relative to a single marker.
In contrast, we base our board on a marker detector with very high accuracy keypoint detection, as we only care about point detection accuracy and identify points on the board using a few low-accuracy ArUco tags.
This leads to a higher number of extracted keypoints and high center point extraction accuracy.

\medskip \noindent \textbf{Invertible Neural Networks.}
Our paper 
models lens distortion using an invertible mapping enforced through the neural network architecture.
Invertible neural networks have been studied extensively in the context of normalizing flows, where network inverses are required for computing log-likelihoods for generative models~\cite{kobyzev2020normalizing, dinh2014nice, chen2018neural, kingma2018glow, pmlr-v97-behrmann19a}.
%
%
Since our application does not require the estimation of the Jacobian for generative tasks, we opt to use an invertible residual network due to its expressive power and convergence speed.
Invertible residual networks have been applied to many tasks, such as shape deformation~\cite{paschalidou2021neural, jiang2020shapeflow, yang2021geometry}, image denoising~\cite{liu2021invertible}, and tone mapping~\cite{mustafa2022distilling}.
In this paper, we explore their applicability to the problem of lens distortion.

\section{Method}
\begin{figure*}[t]
\begin{center}
\includegraphics[width=\linewidth]{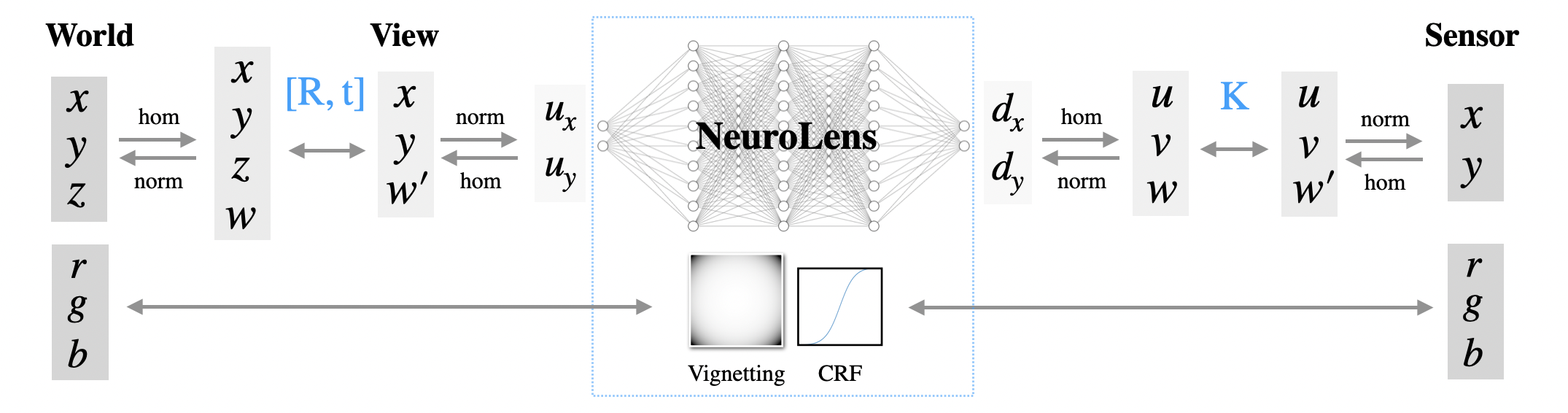}
\end{center}
\vspace*{-7mm}
\caption{\textbf{Mapping Overview.} We illustrate the mapping between three systems: \textbf{World} is the world coordinate/color system, \textbf{View} is the local camera system, \textbf{Sensor} is the sensor coordinate/color system. $hom$ and $norm$ refer to homogenization and de-homogenization operations, respectively. The figure can be read left-to-right to follow a \textbf{projection} operation, right-to-left to follow a \textbf{ray casting} operation.
}
\label{fig:method}
\vspace{-4mm}
\end{figure*}

The goal of camera calibration is to recover the optimal parameters 
that describe the camera model at hand given a set of observations.
The camera model describes the mapping between points $\mathbf{X} \in \mathbb{R}^3$ in the 3D world and their 2D locations $\mathbf{x} \in \mathbb{R}^2$ on the camera sensor.
In this paper, we assume the \emph{projection} component of this mapping to be described by the pinhole camera model.
Under this model, the 2D pixel coordinate $x$ can be obtained by:
\begin{align}
\mathbf{x} = \mathcal{C}(\mathbf{X}) = \operatorname{norm}(\mathbf{K}\cdot(\mathbf{R\cdot X} + \mathbf{t})),\label{eq:simple-camera-model}
\end{align}
where $\operatorname{norm}(\mathbf{x})=(\mathbf{x}[0]/\mathbf{x}[2], \mathbf{x}[1]/\mathbf{x}[2])$, $\mathbf{R}$ and $\mathbf{t}$ are the rotation matrix and translation vector in world-to-camera format, and $\mathbf{K}$ is the intrinsics matrix.
%

This pinhole model, however, captures only some aspects of the true mapping function for real-world cameras: it assumes that light follows a straight line from the world directly to the sensor plane.
This is not the case for real cameras: the optical stack consists of (multiple) lenses with often complex optical properties (\eg fisheye and catadioptric lenses with wide fields-of-view) that cause visible curvature in the projection of straight lines---the \emph{distortion} component of the mapping.
As illustrated in Fig.~\ref{fig:method}, this non-linear distortion can be modeled by a diffeomorphic function $\mathcal{D}$ that maps ideal coordinates $(u_x,u_y)$ to distorted coordinates $(d_x,d_y)$. 
As illustrated in Fig.~\ref{fig:method}, let $u=(u_x,u_y)$ be the normalized coordinates obtained after perspective division but before rescaling by camera intrinsics, the observed pixel coordinate can be obtained by:
\begin{align}
\mathbf{x} = \operatorname{norm}(\mathbf{K}\cdot\operatorname{hom}(\mathcal{D}(u)))\label{eq:camera-model}
\end{align}
In contrast to \eqref{eq:simple-camera-model} which contains a handful of parameters, our camera model $\mathcal{C}$ contains a bijection which is much more complex to model and $\D$ depends on the physical properties of the camera optics. 
Hence, it is important to strike a balance between models with sufficiently many parameters that are at the same time constrained to meaningful lens mappings.
In our work, we propose to model $\D$ using invertible residual networks and show that they are a strikingly simple, well-suited class of functions for modeling distortions. 
Using such functions retains the ability to propagate gradients in either the projection (forward) or casting (backward) operation, enabling end-to-end optimization of the camera intrinsics $\mathbf{K}$, extrinsics $\mathbf{R, T}$, and the distortion mapping $\mathcal{D}$.
%


In what follows, we will first explain how we parameterize the distortion mapping $\mathcal{D}$ using invertible residual networks in Sec.~\ref{sec:camera}.
Then, we develop our training objectives in Sec.~\ref{sec:losses}, in which we consider a common lens effect of vignetting and incorporate the camera response function (CRF) into our model. Finally, we will describe how to obtain keypoints and their corresponding 3D positions for the case of marker-based calibration. In particular, we propose a new pattern that enables very accurate keypoint detection in Sec.~\ref{sec:patterns}.


\subsection{Camera Distortion Model} \label{sec:camera}

%
%
As illustrated in Fig.~\ref{fig:method}, the camera distortion model can be defined as a transformation of a ray from undistorted to distorted directions: $(d_x, d_y) = \D (u_x, u_y)$. In this section, we describe how to parameterize the distortion mapping $\mathcal{D}$. 

Intuitively, the distortion transformation can be used in both directions; therefore the process should be \emph{invertible}.
%
Hence we propose to represent the non-linear distortion as a \emph{diffeomorphism}. 
We can write $\D$ as an invertible function $\D: \Rdom \xrightarrow{} \Rdom$, where its backward mapping $\D^{-1}$ models the undistortion process.
%
We find that invertible neural networks are a suitable model class for regularizing $\D$ as a smooth, invertible function.
Invertible neural networks (INNs) are function approximators that effectively learn differentiable bijections.
Networks that are invertible by construction offer a useful advantage: we can train them on a forward mapping and can use the inverse function at no additional cost. 

Specifically, we propose to parameterize the distortion mapping $\mathcal{D}$ using Invertible Residual Networks (ResNets), a subclass of INNs introduced by Behrmann \etal~\cite{pmlr-v97-behrmann19a}.
Invertible ResNets are composed of residual blocks of the form $f_\theta(x) = x + g_\theta(x)$, where $\theta$ denotes all trainable parameters.
Behrmann \etal show that $f_\theta$ is invertible if $g_\theta$ is Lipschitz-bounded by $1$.
In that case, the inverse of $f_\theta$ can be obtained by computing the fixed-point of function $h(x) = y - g_\theta(x)$, where $y$ is the output of $f_\theta(x)$. 
The fixed-point can be obtained by using the iterative algorithm: $x \leftarrow y - g_\theta(x)$.
In practice, we found that a network with width 1024 and four residual blocks is sufficient. 
For more implementation details, please refer to the supplementary material.

%
%
%

\subsection{Optimization Objectives} \label{sec:losses}

Since we model $\D$ as a differentiable function, it can be used in many optimization scenarios.
For instance, it can be used to optimize keypoints and their corresponding 3D positions obtained using calibration targets introduced in Sec.~\ref{sec:patterns}, or to optimize all camera parameters together with world model parameters during 3D reconstruction.
%

\paragraph{Geometric loss.}  \label{sec:sparse}

A calibration board contains $N$ reference points $\mathbf{X}_i$ whose 3D coordinates are known (in practice, initial estimates for their 3D position can be found using, for example, \cite{extintinit}).
The points are assumed to lie in the $XY$-plane, \ie, their $Z$-component is zero. Given a set of 3D-2D points pairs $\{\mathbf{X}_i, \mathbf{x}_i\}_{i=1}^n$, where $\mathbf{x}_i$ is the detected keypoint position in the image, we can minimize the per-view reprojection error:
\begin{align}
\mathcal{L}_\text{prj}(\Theta) = \|\mathcal{C}_\Theta(\mathbf{X}_i) - \mathbf{x}_i\|_2^2, \label{eq:reproj}
\end{align}
where $\Theta$ includes the camera intrinsics $\mathbf{K}$, extrinsics $\mathbf{R, T}$, and parameters $\theta$ that define the distortion mapping $\mathcal{D}_\theta$.
%
To improve robustness to outliers, we can optionally apply iterative reweighting to Eq.~\ref{eq:reproj} during optimization. 

\medskip \noindent \textbf{Photometric Loss.} \label{sec:dense}
Our model can also be optimized using the gradients from a set of 2D image observations, for example as part of a 3D reconstruction.
If the camera model describes the image formation procedure correctly, then the color at the pixel location predicted by $\mathcal{C}_\Theta(\mathbf{\mathbf{X}})$ should match the color of the actual 3D point $\mathbf{X}$ projected there (assuming constant lighting, exposure and a Lambertian marker material).
Suppose $L(\mathbf{X}) \in \mathbb{R}^3$ is the reference color from the calibration board at 3D location $\mathbf{X}$ and $I_i$ is the color of the observed image at pixel location $\mathbf{x}$.
Their $\ell_2$ difference can be described by $||L(\mathbf{X}) - I_i(\mathcal{C}_\Theta(\mathbf{X}))\|_2^2$.

However, this comparison does not take into account optical effects that influence the mapping from radiance to the final image color.
Most prominently, we also need to model and estimate vignetting effects (radial falloff) present in many zoom and wide angle lenses~\cite{kim2008robust}.
Furthermore, we should take into account the camera response function (CRF), the relationship between the radiance captured by the camera and the resulting sensor readout~\cite{debevec2008recovering}.

To account for such optical effects, we define a function $M(\mathbf{x}, \mathbf{c})=f(V(\mathbf{x}, \mathbf{c}))$ which takes a pixel location $\mathbf{x}$ and the incident radiance $\mathbf{c}$ and returns a sensed color taking into account CRF $f$ and vignetting effect $V$.
Specifically, we parameterize the vignetting function $V$ by $V_\gamma(\mathbf{x}, \mathbf{c}) = \mathbf{c}\cdot\sigma(\operatorname{interp}(\mathbf{x}, \gamma))$, where $\gamma\in\mathbb{R}^{H\times W}$, $\operatorname{interp}$ is bilinear interpolation, and $\sigma$ is a sigmoid function.
In our case, we used a fixed CRF $f$ that is known and uniform across the spatial dimensions of the image.
We include $\gamma$ as part of the camera parameters $\Theta$, which will be jointly optimized. More general formulations can be used for more complex camera response function and vignetting parameterizations that are appropriate for the camera. 
%
%
%
%
%

Finally, if keypoints with known radiance are available, for example from the calibration board described in Sec.~\ref{sec:patterns}, the photometric loss can be used to match the sensed colors to match their expected values.
Given $n$ images $\{I_i\}_{i=1}^n$, we can sample color from $m$ points on the calibration board $\{(\mathbf{X}_j, L_j)\}_{j=1}^m$, and define the photometric loss as:
\begin{align}
\mathcal{L}_\text{pho}(\Theta)=\sum_{i,j} \|M(\mathcal{C}_\Theta(\mathbf{X}_j), L_j) - I_i(\mathcal{C}_\Theta(\mathbf{X}_j))\|_2^2.
\end{align}
%
%
Alternatively, $M$ can be used to map radiance values to color while $C_\Theta$ are the rays cast for a gradient-based optimization of a radiance field---in that case the gradients can be naturally used to update all relevant parameters (see Sec.~\ref{sec:neural_render}).
%

\subsection{Marker-based Calibration}\label{sec:patterns}

%
%
%
%
The most common optimization scenario for camera calibration uses an established set of corresponding keypoints to determine $\D$.
These are often obtained using a calibration board with a known marker structure that allows for identifying keypoints with high precision.
The classical OpenCV~\cite{opencv_library} library, as well as more recent methods~\cite{schops2020having,lochman2021babelcalib} use different calibration board types to achieve this.
All these board types are hand-designed: their respective patterns yield points with high contrast that can be readily identified.
Still, it is not trivial to achieve sub-pixel accurate keypoint detections. 
In particular, checkerboard corner detection utilizes line-fitting to identify intersection points, and star-shaped pattern detectors rely on symmetric features to identify keypoint centers.
All these strategies are non-trivial to implement and are adversely affected by lens distortion.

To address this problem, we propose to optimize the keypoint marker design together with a deep-learning based keypoint detector end-to-end.
To represent markers, we create a three-channel tensor that stores an RGB image.
To optimize it, we create a simplified model of the image formation process from the marker definition that can contain: printing (small local distortions), lighting (slight intensity changes), motion blur, perspective distortion (viewpoint changes), lens distortion, and color aberration.
In our experiments, we use a single marker design optimized for fairly general use by implementing some of the aforementioned effects using blurring, affine transformations, added noise and color distortion.
The detector, a MobileNet-v3~\cite{mobilenetv3} with a simplified 2D location prediction optimized using a Gaussian negative log likelihood of the true keypoint location, has the task of localizing the marker center. 
This means, we use a fully supervised training for the entire detection process that can be adjusted to match the capture scenario at hand.

\begin{figure}[t]
\begin{center}
\includegraphics[width=\linewidth]{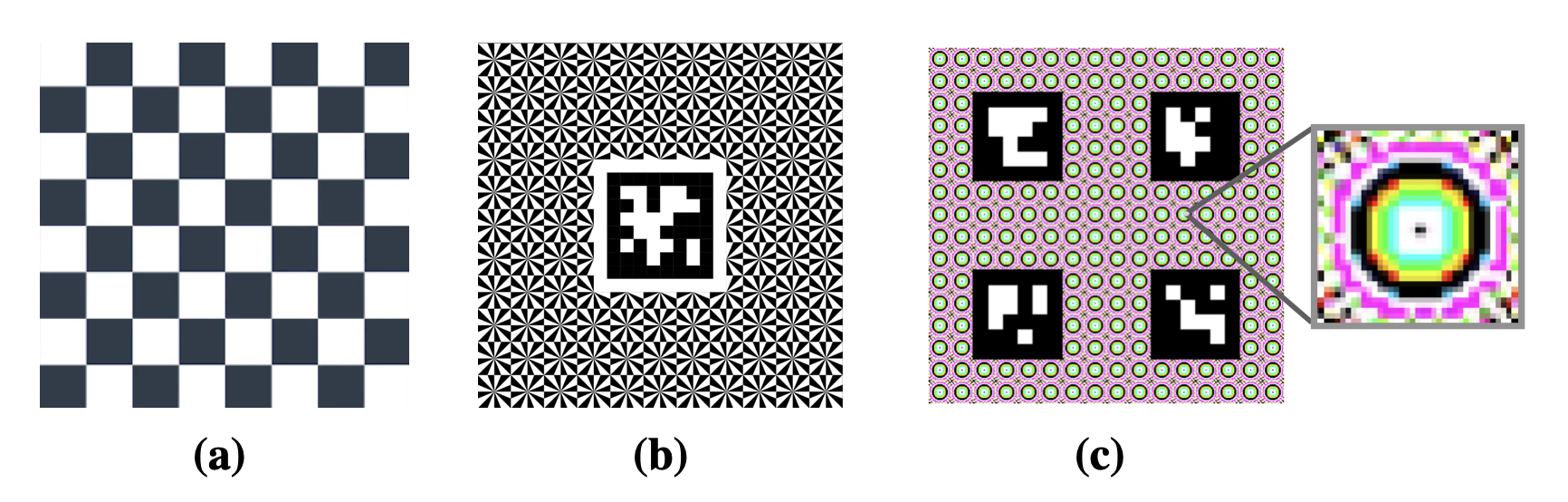}
\end{center}
\vspace*{-7mm}
\caption{
\textbf{Keypoint Patterns \& Markers.} \textbf{(a)} Checkerboard pattern, \textbf{(b)} Star-shaped pattern proposed in \cite{schops2020having}. \textbf{(c)} Our proposed calibration pattern, allowing for unique localization using ArUco tags~\cite{aruco1Garrido-Jurado2016,aruco2Romero-Ramirez2018}, and containing high-contrast patterns for accurate keypoint detection. The markers can be optimized specifically for the camera and capture scenario; the size and ratio of markers and tags can be adapted according to the resolution of the camera.}
\label{fig:markers}
\vspace{-5mm}
\end{figure}

The result is empirically superior to other marker shapes and makes better use of color (as shown in Tab.~\ref{tab:table_eval}): 
in contrast to the black and white patterns used in manual marker design like checkerboards, our machine-optimized markers use color to maximize cues about 
keypoint location (see Fig.~\ref{fig:markers}).
The symmetry of the marker emerges from the optimization to achieve rotation invariance.
%
%
The center keypoint is marked black with a small white area around it to maximize contrast and be robust to color bleeding; several circles around it provide additional information to identify and localize it.
A pattern board can be readily assembled using these markers by using ArUco~\cite{aruco1Garrido-Jurado2016,aruco2Romero-Ramirez2018} markers to identify planar areas and rough sizes, extracting candidate areas and running the pre-trained detector to obtain marker locations.
In practice, the confidence prediction from the predicted Gaussian variance helps to filter uncertain detections.
Thanks to the high efficiency of MobileNet-v3, the detector runs at multiple frames per second allowing live data acquisition feedback.

\section{The SynLens Dataset}
Evaluating lens models is inherently hard: ground truth is nearly impossible to obtain (since it would require a possibly destructive analysis of equipment), and performing measurements at scale requires a large supply of cameras and lenses.
On the other hand, over the last years the LensFun database\footnote{\url{https://lensfun.github.io/}} has steadily grown and accumulated a large set of crowd-sourced high-quality measurements of lens characteristics.
Hence, we propose to use it to create a large dataset of high quality \emph{synthetic} lenses that can be used to evaluate calibration models.
By creating the data synthetically, we can perform calibration in perfect control of noise characteristics and create informative estimates of calibration performance on many consumer devices.

\medskip \noindent \textbf{The Data.}
%
%
The LensFun database contains more than 3,500 lens models from 40 different camera makers, \eg Canon, Nikon, action cams, \etc. 
For each lens profile, it specifies lens model, focal length, lens distortion, vignetting and chromatic aberration (TCA).
High-quality data was collected by photography enthusiasts using the open-source Hugin software.
Of this data, we selected 400 lenses by choosing 10 different lens types for each camera maker.
%

Using this data, we offer dataset users an API to render images, and specifically calibration boards, through these lenses while automatically applying $\D$ and $V$.
To test calibration specifically, we provide options to obtain the ground truth positions of projected keypoints.
In the following, we describe the API functionality.

%
\medskip \noindent \textbf{Virtual camera set-up.}
We deploy a virtual perspective camera in a synthetic scene using PyTorch3D~\cite{ravi2020pytorch3d}. 
It is easy to adjust the virtual camera parameters and to control its pose. 
We point the camera at a calibration target using several in-plane rotation $\alpha$ and zenith $\beta$ angles.
For each scene, we translate the camera off-center and obtain a series of 200 non-parallel images at resolution of $1024\times1024$. 

\medskip \noindent \textbf{Lens distortion.}
%
%
\begin{figure}[t]
\begin{center}
\includegraphics[width=\linewidth]{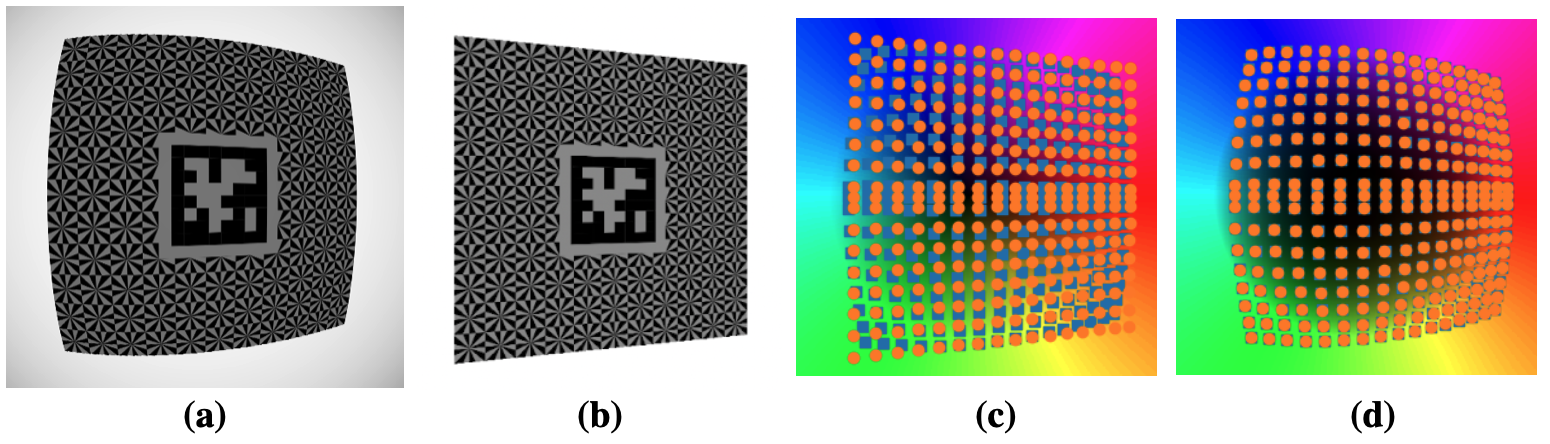}
\end{center}
\vspace*{-5mm}
\caption{\textbf{An example from the SynLens Dataset}: \textbf{(a)}~distorted frame, \textbf{(b)}~corresponding undistorted and normalized view, \textbf{(c)}~initialization of keypoints and \textbf{(d)}~keypoints after optimization. \textbf{Blue:}~ground truth keypoint positions, \textbf{orange:}~predicted keypoint positions. \textbf{Hue}: offset direction, \textbf{saturation}: offset magnitude. 
}
\label{fig:synlens}
\end{figure}
\begin{table}[tb]
  \centering
    \begin{tabular}[t]{lcc}
        \toprule
          Models & Formulation ($C$) \\
         \midrule
          Poly3  & $r_d = r_u (1 - k_1 + k_1 r_u^2) $  \\
          Poly5 &  $r_d = r_u (1 + k_1 r_u^2 + k_2 r_u^4) $  \\
          PTLens & $r_d = r_u (a r_u^3 + b r_u^2 + c r_u +  1 - a - b - c) $ \\
        \bottomrule
    \end{tabular}
    \vspace*{-2mm}
    \caption{\textbf{Analytic equations in LensFun.}}
  \label{tab:table_models}
  \vspace{-5mm}
\end{table}
From the Lensfun database, lens distortion information is available in one of several predefined formats: PTLens, poly3, poly5 or Adobe Lens (see Tab.~\ref{tab:table_models}).
%
%
%
%
%
%
%
According to each calibrated lens model in Lensfun, we synthetically generate distorted and undistorted point pairs in the normalized image domain.
%
%

\medskip \noindent \textbf{Vignetting.}
%
The vignette function in the database is parameterized as the polynomial radial loss function
$V(r_d) = 1 + k_1 r_d^2 + k_2 r_d^4 + k_3 r_d ^ 6,$
where $k1$, $k2$, $k3$ are a set of vignetting parameters; these model parameters are identical for all color channels.
%
%
An example of a simulated lens recording a calibration pattern from~\cite{schops2020having} including distortion and vignetting is shown in Fig.~\ref{fig:synlens}. 
The vignetting effects are clearly visible in (a), whereas (b) shows a successful calibration result.
We show recorded and optimized keypoints as well as a visualization of the lens model in subfigures (c) and (d).

\section{Experiments}\label{sec:experiments}

In our experiments, we compare the performance of our lens models and marker board on the proposed SynLens dataset with several established methods before presenting results on real-world data for keypoint-based calibration and radiance-field reconstruction on radial and fisheye lenses.
%

\subsection{Evaluation on SynLens} \label{sec:evaluate_synthetic}
\begin{table}[htbp]
  \centering
    \begin{tabular}[t]{lcccc}
        \toprule
        \multicolumn{1}{c}{} & \multicolumn{3}{c}{Camera Models} & \multicolumn{1}{c}{} \\
        \cmidrule(rl){2-4} 
          Methods & {Poly3} & {Poly5} & {PTLens} & {Avg} \\
        \cmidrule(r){1-1} \cmidrule(rl){2-5}
          Sch\"ops \etal~\cite{schops2020having} & 0.162 & 0.124 & 0.121 & 0.135  \\
          Ours   & 0.104 & 0.052 & 0.061 & 0.072  \\
        \bottomrule
    \end{tabular}
    \vspace*{-2mm}
    \caption{\textbf{Reprojection error (RMS) on SynLens by method and lens model for ground truth keypoints.}}
    \label{tab:gt}
    \begin{tabular}[t]{lccc}
        \toprule
        \multicolumn{1}{c}{} & \multicolumn{3}{c}{Keypoint Types} \\
        \cmidrule(rl){2-4} 
          Methods & {Checkboard} & {Star} & {Ours}\\
        \cmidrule(r){1-1} \cmidrule(rl){2-4}
          OpenCV~\cite{opencv_library} & 0.152 & 0.175 & 0.129  \\
          Sch\"ops \etal~\cite{schops2020having} & 0.178 & 0.141 & 0.158  \\
          Ours   & 0.154 & 0.130 & 0.114  \\
        \bottomrule
    \end{tabular}
    \vspace*{-2mm}
    \caption{\textbf{Reprojection error (RMS) on SynLens for detected keypoints by model and keypoint type.}}
    \label{tab:table_eval}
\end{table}


%
On SynLens, we establish baseline comparisons with two widely used camera calibration methods and board patterns: (1)~the distortion model implemented in OpenCV~\cite{opencv_library} using all distortion terms, and the board and method from Sch\"ops \etal~\cite{schops2020having}, a state-of-the-art generic calibration method with an open-source implementation. 
Since we evaluate on a synthetic dataset, the ground-truth locations of keypoints can be obtained by transforming them using Eq.~\ref{eq:camera-model}.
We optimize a lens model for each camera lens using the keypoint correspondences; once using the ground-truth project locations of the keypoints, once by using the respective keypoint detector.
We then measure the root-mean-squared (RMS) reprojection error (\ref{eq:reproj}) on a set of 20 held-out test images uniformly sampled from each sequence.

Tab.~\ref{tab:gt} shows a breakdown of calibration performance for \cite{schops2020having} and our method for the different camera models in the dataset for ground truth keypoint projections.
We do not use OpenCV in this table since OpenCV uses the exact same parameterization for its model as has been used to generate the data, and therefore unsurprisingly achieves perfect fitting in this scenario.
Our method outperforms Sch\"ops \etal in this setting for all camera models by a large margin, even though Sch\"ops \etal also use a highly parameterized model.

Tab.~\ref{tab:table_eval} shows the performance of all methods, using different calibration targets and detection results from the respective keypoint detectors.
%
%
We expect the best results on the diagonal of the table (each method performing best with its own type of detector pattern and keypoint detector).
This mostly holds true, except for OpenCV does better with out marker than with the checkerboard.
Our proposed method achieves the overall lowest RMS error with the proposed calibration target in this setting.
We analyze how different levels of artificial keypoint noise as well as the severity of the distortion affects the calibration performance of different methods in the supplemental material.
\begin{figure}[t]
\begin{center}
\includegraphics[width=\linewidth]{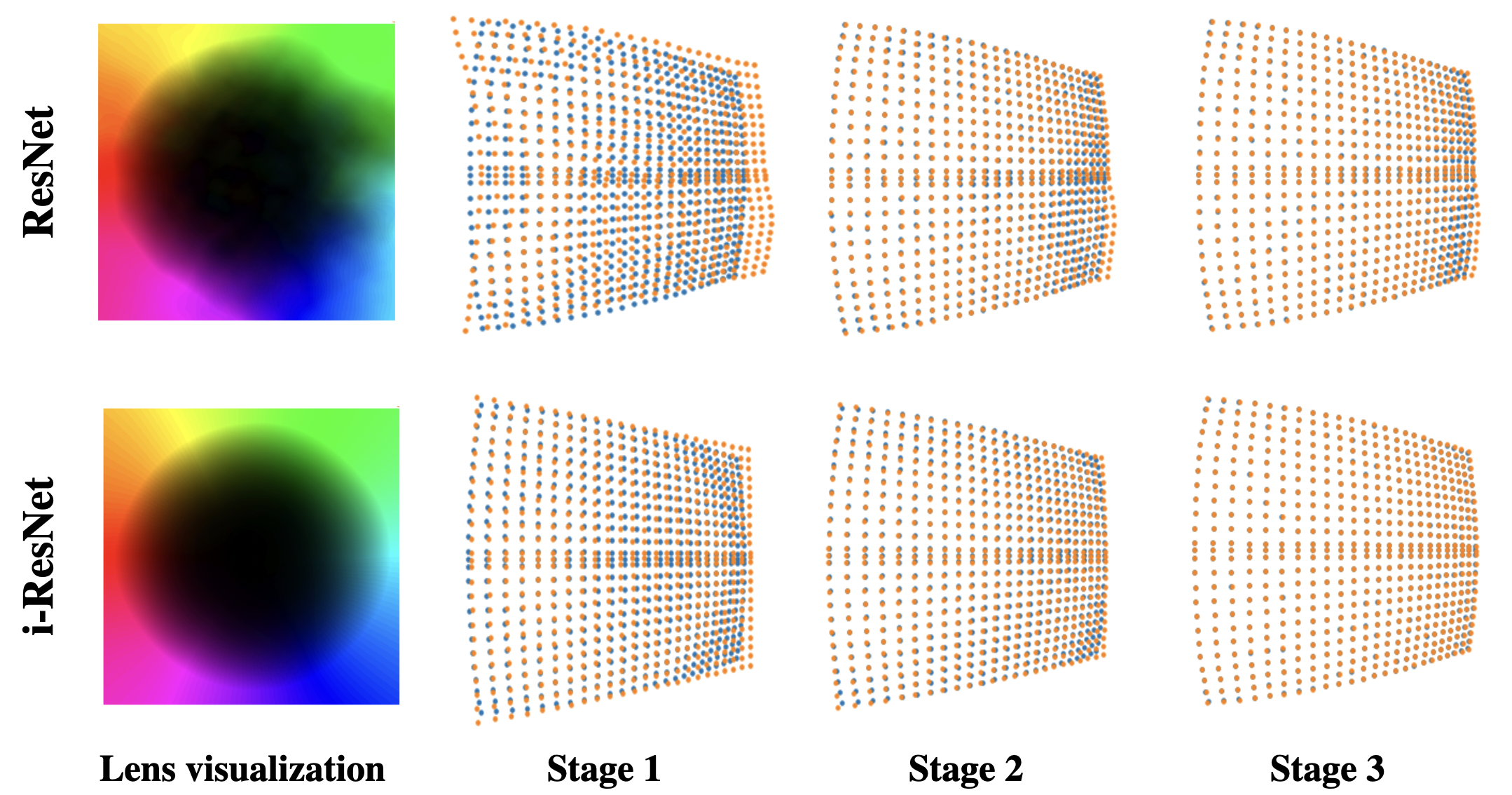}
\end{center}
\vspace{-6mm}
\caption{
\textbf{Comparison between ResNet and invertible ResNet lens models.} Left to right: optimized lens model visualization, and three stages during training, for each: ground-truth keypoints (blue) and projected keypoints (orange) on test images. The rainbow visualization is described in Fig.~\ref{fig:synlens}.
}
\label{fig:resnet}
\vspace{-5mm}
\end{figure}
\vspace*{-0mm}
\begin{table}[htbp]

  \centering
    \begin{tabular}[t]{lcccc}
        \toprule
        \multicolumn{1}{c}{} & \multicolumn{2}{c}{i-ResNet} & \multicolumn{2}{c}{ResNet}  \\
        \cmidrule(r){2-3} \cmidrule(rl){4-5}
          Ratio (tr : val) & {Train} & {Val} & {Train} & {Val} \\
         \cmidrule(rl){1-5}
          1 : 1  & 0.16 &	0.16 &	0.30 & 0.45  \\
          1 : 4  & 0.15 &	0.28 &	0.32 &	0.60  \\
          1 : 8  & 0.15 &	0.33 &	0.37 &	1.56 \\
        \bottomrule
    \end{tabular}
    \vspace*{-2mm}
    \caption{\textbf{RMS error comparison between ResNet and i-ResNet for different training set sizes.} Total number of keypoints in the validation set remains the same across all experiments.}
  \label{tab:table_resnet}
\end{table}
\subsection{Comparison with ResNet} \label{sec:resnet}
The Lipschitz constraint on the invertible ResNet is a powerful regularizer for the proposed model.
%
%
%
Compared with standard ResNets, we find that invertible ResNets are less likely to be affected by outliers because they are implicitly constrained to model a smooth function.
In Fig.~\ref{fig:resnet}, we show a comparison of a ResNet and invertible ResNet trained on a lens with noisy keypoint detections.
The ResNet overfits to the noisy measurements present in the training data, for example at the top left corner.
In comparison, the invertible ResNet can model accurate lens geometry ofand makes continuous progress towards a reasonable solution over the course of the optimization.
In Tab.~\ref{tab:table_resnet}, we show that invertible ResNets are robust to reduced amounts of supervision thanks to their stronger priors. 

\begin{figure}[t]
\begin{center}
\includegraphics[width=\linewidth]{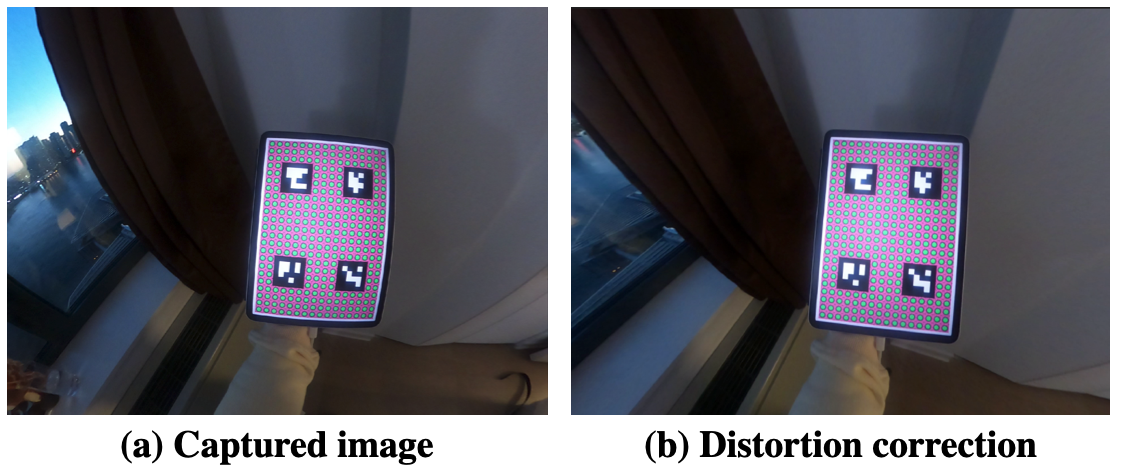}
\end{center}
\vspace*{-0.6cm}
\caption{
\textbf{Undistortion of a GoPro super-wide recording.}
}
\label{fig:gopro}
\end{figure}

\begin{figure}[t]
\begin{center}
\includegraphics[width=\linewidth]{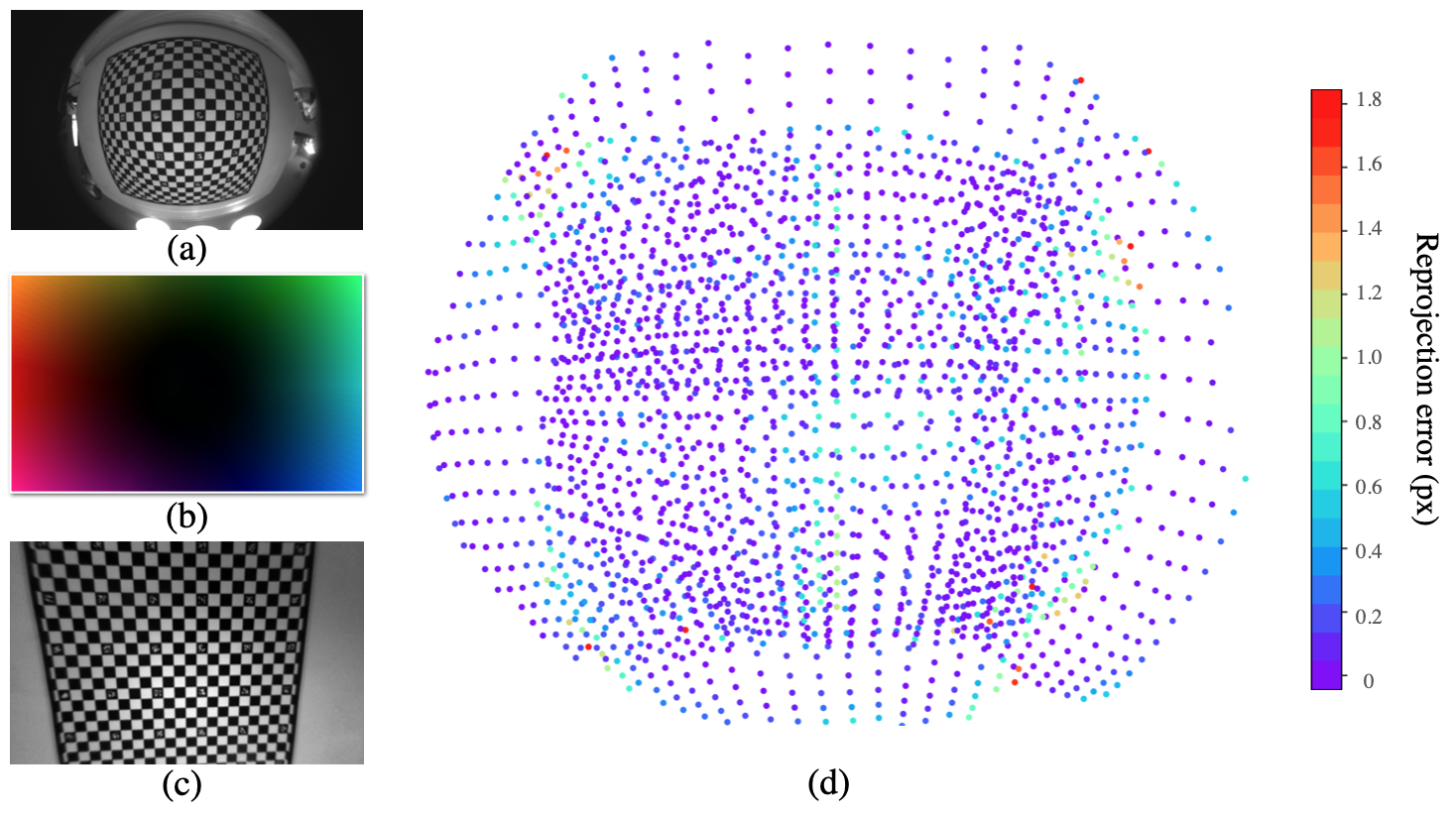}
\end{center}
\vspace*{-6mm}
\caption{
\textbf{OCamCalib Fisheye camera calibration.} \textbf{(a)}~example frame captured by the Fisheye camera, \textbf{(b)}~lens distortion map (hue: distortion direction; saturation: distortion magnitude), \textbf{(c)}~undistorted image, \textbf{(d)}~residuals of reprojected keypoints on test images.
}
\label{fig:real}
\vspace{-4mm}
\end{figure}
\begin{figure*}[!htb]
    \centering
    \vspace*{-2mm}
    \includegraphics[width=0.98\textwidth]{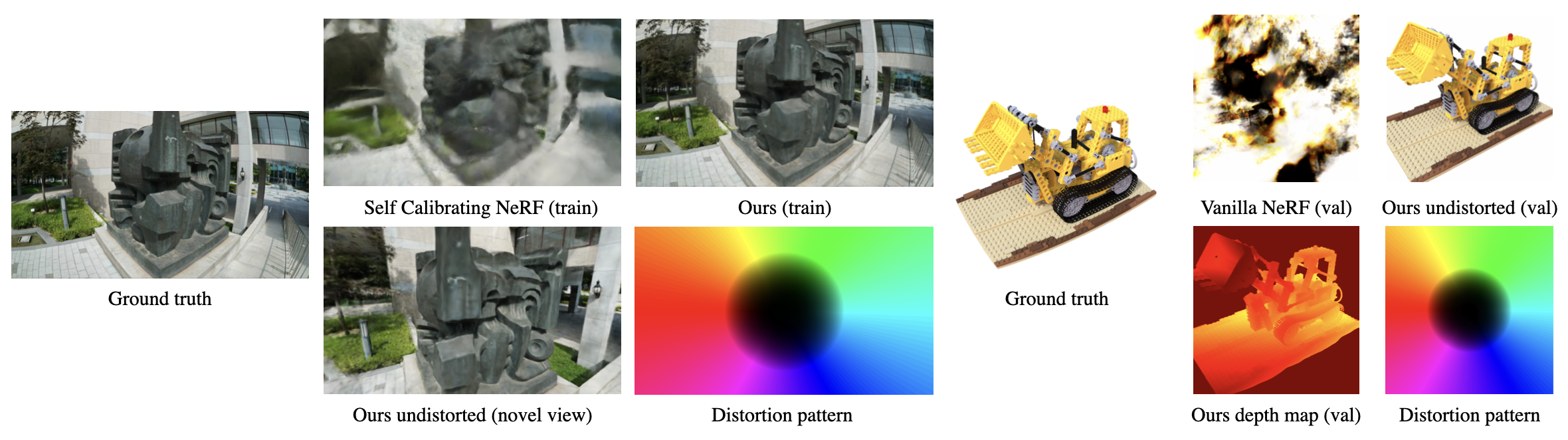}
    \vspace*{-4mm}
    \caption{\textbf{Neural Radiance Field results.} Qualitative results on the FisheyeNeRF~\cite{jeong2021self} and original NeRF datasets. \textbf{Left:} the FisheyeNeRF dataset stretches the capabilities of in-the-wild calibration without keypoint correspondence to the limits. Baseline method SC-NeRF~\cite{jeong2021self}, to the best of our knowledge, only shows results on training views, for which our model fares remarkably well. \textbf{Right:} Results on a scene from the NeRF dataset. We used a Blender scene and added significant barrel distortion (as visible in the ``Ground truth" setting). NeRF~\cite{mildenhall2020nerf} fails to reconstruct the scene. Our method manages to retrieve the lens parameters well, resulting in high-quality reconstruction and depth.}
    \label{fig:neural_rendering_comparison}
\vspace{-4mm}
\end{figure*}


\subsection{Evaluation on Real Captures} \label{sec:evaluate_real}

To ensure that our evaluation results on synthetic data carry over to real-world capture scenarios, we conduct several experiments using challenging wide angle and fisheye lenses.
%
%
In the first experiment we attempt calibration for a consumer GoPro camera with wide and super-wide lens settings.
For data collection, we captured a video of a board with our proposed calibration pattern.
We then run keypoint detection and fit our lens model to each camera.
Fig.~\ref{fig:gopro} shows the undistortion result for the super-wide lens.
For both lenses we achieve slightly better result on super-wide lens setting than OpenCV on held-out test frames:  RMS score of OCV 1.50 vs.\ Ours 1.46, while having comparable results on wide lens settings, OCV 0.56 vs.\ Ours 0.61.
%
%
%
%
%

In the second setting, we extend our experiment to a very challenging scenario: the OCamCalib~\cite{scaramuzza2006flexible} dataset, with camera field of view ranging from 130$^{\circ}$ to 266$^{\circ}$ and the UZH~\cite{delmerico2019we} dataset, which consists of eight wide-angle and fisheye cameras with fields of view ranging from 124$^{\circ}$ to 166$^{\circ}$.
%
%
%
Keypoint detections are available from a planar chessboard target marked with AprilTags.
%
%
%
We compare our results with the state-of-the-art camera calibration framework BabelCalib~\cite{lochman2021babelcalib}. As shown in Tab.~\ref{tab:table_babelcalib}, our method outperforms BabelCalib on most cameras from the UZH dataset. We also visualize the residuals of the reprojected keypoints of test images in Fig.~\ref{fig:real} from OCamCalib, on which our method achieves an \emph{unweighted} reprojection error of 0.91 (all points contribute equally to the error metric). This is a comparable score with the BabelCalib system with a significantly simpler model on this challenging data.

\begin{table}[htbp]
  \centering
    \resizebox{0.48\textwidth}{!}{
    \begin{tabular}[t]{lccccccccc}
        \toprule

        \multicolumn{1}{c}{} & \multicolumn{4}{c}{UZH-DAVIS} & \multicolumn{4}{c}{UZH-Snapdragon} &  \\
        
        \cmidrule(rl){2-5}  \cmidrule(rl){6-9}
          Methods & {$\operatorname{I-1}$} & {$\operatorname{I-2}$}  & {$\operatorname{O-1}$}  &  {$\operatorname{O-2}$}  & {$\operatorname{I-1}$} & {$\operatorname{I-2}$}  & {$\operatorname{O-1}$}  &  {$\operatorname{O-2}$}   & Mean \\
        \midrule
          BabelCalib & 0.31 & 0.69 & 1.58 & 0.44& \bf{0.28} & 1.10 & 0.68 & \bf{0.28} & 0.67\\
          Ours  & \bf{0.25} & \bf{0.52} & \bf{0.49} & \bf{0.36}  &  0.64 & \bf{0.57} & \bf{0.34} & 1.08 & \bf{0.53}\\
        \bottomrule
    \end{tabular}
    }
    \vspace*{-2mm}
    \caption{\textbf{RMS score comparison between BabelCalib and our method on UZH camera dataset.}}
      \label{tab:table_babelcalib}
      \vspace{-5mm}
\end{table}

%
%
%

\subsection{Neural Radiance Fields} \label{sec:neural_render}

A significant advantage of our proposed lens model is its two-way differentiability, making it simple to deploy in 3D reconstruction workflows.
Neural radiance fields (NeRFs)~\cite{mildenhall2020nerf} are a state-of-the-art approach for novel-view synthesis.
They optimize a scene model directly from RGB images, given the camera intrinsics and poses.
While NeRF achieves high-quality novel views, it requires accurate camera parameters, which can be difficult to obtain in practice, particular for lens parameters, which often require an additional calibration stage.
We integrate our neural lens model into a neural rendering framework~\cite{muller2022instant} such that the camera poses, pinhole intrinsics and lens distortion are optimized together with the appearance model, given only RGB observations.
The camera intrinsic and extrinsic parameters are initialized using values obtained from a photogrammetry software package, Metashape, yet undistorted.
As can be seen in Fig.~\ref{fig:neural_rendering_comparison}, our approach achieves a high-quality representation of camera views and successfully recovers the lens distortion, even in the case of extremely distorted recordings from the FisheyeNeRF dataset~\cite{jeong2021self}.
%

To experiment with significant distortion, but still in a non-fisheye setting, we use the NeRF dataset and augment a Blender scene with barrel distortion.
The NeRF reconstruction fails completely in this setting; augmented with our proposed model it succeeds in reconstruction and undistortion without any other changes to the training pipeline.
\section{Limitations and Future Work}
%
%
In cases of very extreme lens distortion, it could be helpful to initialize the model with a prior expectation as opposed to starting from an identity initialization.
This could help the convergence rate as well as lead to even better solutions.
Incorporating lens priors for specific models could also be used for model regularization if that's desired for the specific application, though we found the proposed model to be very stable and usually not needing additional regularizers.
%
%
%
%


\section{Conclusion}
In this paper, we presented a novel approach for neural lens modelling with a focus on end-to-end optimization, generality and ease-of-use in existing deep learning pipelines.
It includes distortion as well as vignetting effects and, thanks to being based on invertible residual network models, can be optimized for projection and raycasting.
The model can directly be used to improve the results for 3D reconstructions for radiance field models with hardly any changes to existing implementations.
We also introduced an end-to-end differentiable marker-board and point detector that can be used to perform offline calibration.
Using our proposed synthetic lens dataset as well as results on GoPro and fisheye camera, we showed that the proposed model generalizes across lenses, cameras and applications and can be a reliable calibration component for future applications and research.

{\small
\bibliographystyle{ieee_fullname}
\bibliography{main}
}

\end{document}